%% file: RA-L-Template-YY.tex
\documentclass[letterpaper, 10 pt, journal, twoside]{IEEEtran}
\ifCLASSINFOpdf
\else
\fi
\usepackage{color}
\usepackage{xspace}
\usepackage{amsmath}
\usepackage{bbm}
\usepackage{dsfont}
\usepackage{amsfonts}
\usepackage{graphicx}
\usepackage{booktabs}
\usepackage{caption}
\usepackage{makecell}

\input{commands}

\begin{document}

\title{
MOLTR: Multiple Object Localisation, Tracking and Reconstruction from Monocular RGB Videos 
}


\author{Kejie Li$^{1}$, Hamid Rezatofighi$^{2}$, and Ian Reid$^{1}$%
\thanks{Manuscript received: Octobor, 15, 2020; Revised January, 10, 2021; Accepted February, 5, 2021.}
\thanks{This paper was recommended for publication by Editor Sven Behnke upon evaluation of the Associate Editor and Reviewers' comments.
This work was supported by the University of Adelaide, Australian Centre for Robotic Vision, and Monash University} 
\thanks{$^{1}$Kejie Li and Ian Reid  are with the School of Computer Science, at the University of Adelaide, and Australian Centre for Robotic Vision
        {\tt\footnotesize kejie.li@adelaide.edu.au}}%
\thanks{$^{2} $Hamid Rezatofighi is with Department of Data Science and AI, Faculty of Information Technology, Monash University,
VIC, Australia.
        {\tt\footnotesize hamid.rezatofighi@monash.edu}}%
\thanks{Digital Object Identifier (DOI): see top of this page.}
}
\markboth{IEEE Robotics and Automation Letters. Preprint Version. Accepted February, 2021}
{LI \MakeLowercase{\textit{et al.}}: MOLTR} 

%



\twocolumn[{%
\renewcommand\twocolumn[1][]{#1}%
\maketitle
\begin{center}
    \centering
    \includegraphics[width=0.95\linewidth]{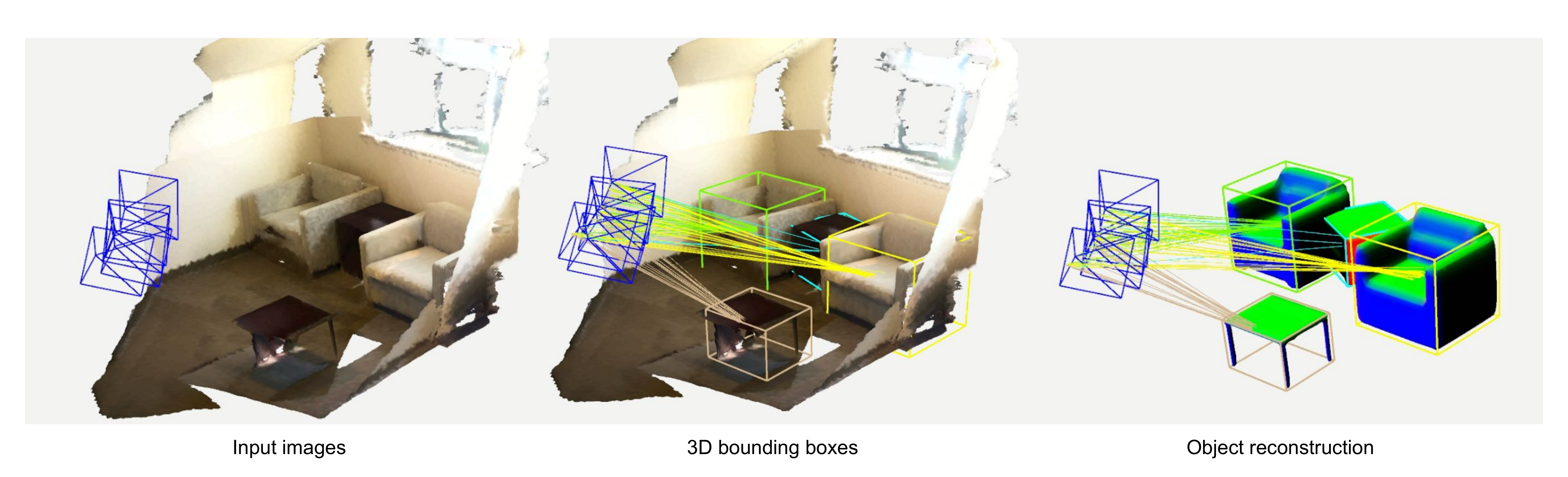}
    \captionof{figure}{\small{A subset of input RGB images are represented by blue frustums at the left image. Detection and tracking is shown on the middle image, where the colored rays indicate the associated detections to the same object. The object-level reconstruction from \methodname is shown on the right image. Note that the scene mesh is \textbf{not} used by \methodname, shown for visualization purpose only.}}
\end{center}%
}]
\renewcommand*{\thefootnote}{\fnsymbol{footnote}}
\footnotetext{Manuscript received: Octobor, 15, 2020; Revised January, 10, 2021; Accepted February, 5, 2021.}
\footnotetext{This paper was recommended for publication by Editor Sven Behnke upon evaluation of the Associate Editor and Reviewers' comments.
This work was supported by the University of Adelaide, Australian Centre for Robotic Vision, and Monash University}
\footnotetext{$^{1}$ Kejie Li and Ian Reid are with the School of Computer Science and the Australian Institute for Machine Learning, at the University of Adelaide, Australia, and also with the Australian Centre for Robotic Vision. {\tt\footnotesize kejie.li@adelaide.edu.au}}
\footnotetext{$^{2}$ Hamid Rezatofighi is with Department of Data Science and AI, Faculty of Information Technology, Monash University,
Clayton, VIC, Australia. {\tt\footnotesize hamid.rezatofighi@monash.edu}}
\footnotetext{Digital Object Identifier (DOI): see top of this page.}
\renewcommand*{\thefootnote}{\arabic{footnote}}
\setcounter{footnote}{0}

\begin{abstract}
Semantic aware reconstruction is more advantageous than geometric-only reconstruction for future robotic and AR/VR applications because it represents not only \emph{where} things are, but also \emph{what} things are.
Object-centric mapping is a task to build an object-level reconstruction where objects are separate and meaningful entities that convey both geometry and semantic information. 
In this paper, we present \methodname, a solution to object-centric mapping using only monocular image sequences and camera poses. 
It is able to localise, track and reconstruct multiple rigid objects in an online fashion when an RGB camera captures a video of the surrounding.
Given a new RGB frame, \methodname firstly applies a monocular 3D detector to localise objects of interest and extract their shape codes representing the object shape in a learnt embedding. Detections are then merged to existing objects in the map after data association. Motion state (\ie kinematics and the motion status) of each object is tracked by a multiple model Bayesian filter and object shape is progressively refined by fusing multiple shape code.
We evaluate localisation, tracking and reconstruction on benchmarking datasets for indoor and outdoor scenes, and show superior performance over previous approaches.
\end{abstract}


%
\IEEEpeerreviewmaketitle

\section{Introduction}
\input{c0_introduction}

%
\begin{figure*}[h]
    \centering
    \includegraphics[width=0.95\linewidth]{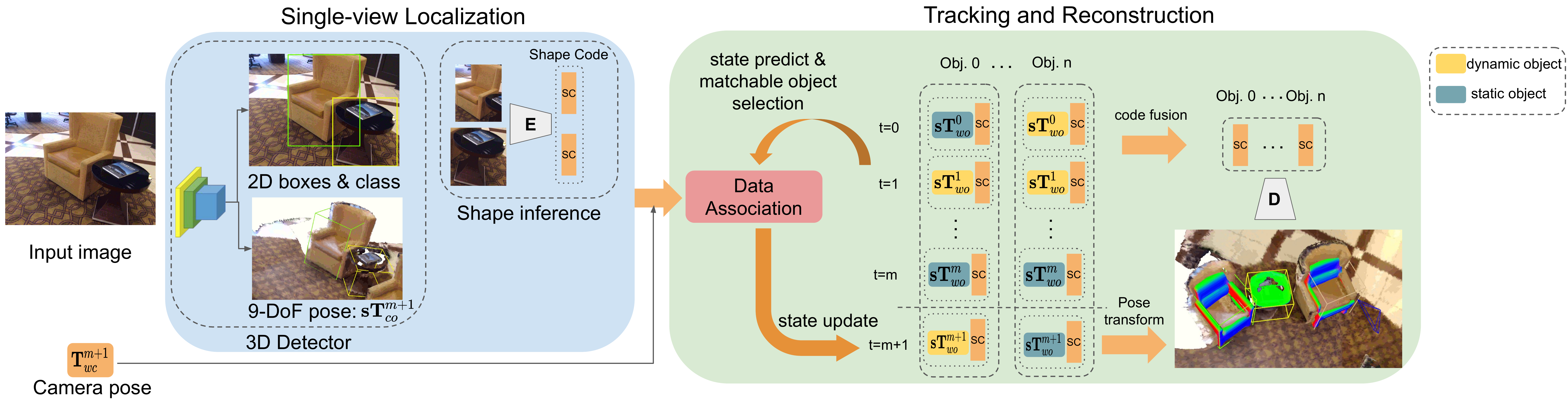}
    \caption{\small{\methodname pipeline. Given a new RGB frame, we predict 6-DoF object pose with respect to the camera $\mathbf{T}_{co}^{m+1}$ and object scale $\mathbf{s}$ (\ie 3D dimension) for each object of interest, which is visualised as an oriented 3D bounding box. We also predict object class and 2D bounding box for each object. An image patch cropped by the 2D bounding box is mapped to a single-view shape code via the shape encoder. The state of objects in the map is tracked by a multiple model Bayesian filter. The motion status is indicated by different background colours of object poses. After filter prediction, matchable objects are used to associate to the new set of detections. A matched detection is attached to the object track, and the shape is progressively reconstructed by decoding the fused shape code.}}
    \label{fig:pipeline}
    \vspace{-0.3cm}
\end{figure*}

\section{Related Work}
\input{c1_related_work}

\section{Method}
\input{c2_method}

\section{Experiments}
\input{c3_experiment}

\section{Conclusion}
\input{c4_conclusion}


%

\section*{Acknowledgment}
We gratefully acknowledge the support of the Australian Research Council through the Centre of Excellence for Robotic Vision CE140100016 and  Laureate Fellowship FL130100102

\ifCLASSOPTIONcaptionsoff
  \newpage
\fi



%


\bibliographystyle{IEEEtran}
\bibliography{IEEEabrv,OurRef}

%

\end{document}

%% file: commands.tex
\newcommand{\ie}{\textit{i.e.}\xspace}
\newcommand{\eg}{\textit{e.g.}\xspace}
\newcommand{\etal}{\textit{et al.}}

\newcommand{\methodname}{\text{MOLTR}\xspace}

\newcommand{\revise}[1]{{#1}}
\newcommand{\fixtypo}[1]{{#1}}

%% file: c0_introduction.tex
\IEEEPARstart{R}{econstructing} the 3D environment from images is a fundamental problem in robotics and computer vision. Early real-time approaches to this problem are sparse SLAM systems~\cite{klein2007parallel, davison2007monoslam} that represent the map as a set of sparse 3D points. The increasing computation power enables dense SLAM systems~\cite{newcombe2011dtam, newcombe2011kinectfusion, whelan2015elasticfusion}, where the reconstruction is composed of dense surfels or Truncated Signed Distance Function (TSDF).

Despite the high-quality geometric reconstruction produced by the aforementioned frameworks, for intelligent robotic applications that need to interact with the environment (\eg fetching objects, tidying up rooms), it is essential to have the knowledge of both geometry and instance-level semantic information of a scene.
With the advance of semantic understanding using deep neural nets, object-centric mapping, where the geometry and semantic properties of the environment are jointly carried out in the form of object instances has gained rapid progress. 

In this paper, we are concerned with the problem of \textit{online} detection, tracking, and reconstruction of \textit{potentially dynamic} \revise{rigid} objects from \textit{monocular} videos. 
Although elements of this problem have been tackled extensively, few works have adequately addressed all three -- online, dynamic, and monocular -- simultaneously. 
Our system \methodname judiciously combines a number of traditional and deep learning techniques to address the problem. 
Given a new RGB frame, a monocular 3D detector is used to localise objects presented in this view, and each detected object is mapped to a learned shape embedding by our shape encoder. 
Meanwhile, the state, which includes kinematics and motion status (\ie dynamic/static), of each existing object in the map is tracked via a multiple model Bayesian filter. 
Matchable objects, selected based on the motion state, are used to associate with the newly detected objects, after which, associated detections are merged to the map, the filters are updated, and object shapes are incrementally refined by shape code fusion.

To summarise, the main contributions of our work are :
\begin{itemize}
\item We present \methodname, a unified framework for object-centric mapping, which is able to localise, track, and reconstruct multiple objects in an online fashion given monocular RGB videos.
\item We demonstrate that the combination of monocular 3D detection, multiple model Bayesian filter and deep learned shape prior leads to robust multiple object tracking and reconstruction.
\item We evaluate the proposed system extensively showing more accurate reconstruction and robust tracking than previous approaches on both indoor and outdoor datasets. 
\end{itemize}

%% file: c1_related_work.tex
In recent years, we have seen impressive progress in semantic aware reconstruction. 
Early works~\cite{stuckler2014multi, hermans2014dense, pham2015hierarchical} use graphical models to assign semantic labels to a geometric reconstruction.
SemanticFusion~\cite{mccormac2017semanticfusion} employs a deep network to predict pixel-wise semantic labels given RGB frames, which are then fused into a semantic mapping by leveraging the geometric reconstruction from an RGBD SLAM. Although these approaches enrich the geometric reconstruction by attaching semantic labels, they are not object-centric as they cannot separate objects of the same class.

Pioneering works on the object-centric mapping are based on template matching and thus is limited to a set of \textit{a-priori} known objects. G{\'a}lvez-L{\'o}pez \etal~\cite{galvez2016real} propose a monocular-based SLAM that matches detections against objects in a database using bags of binary words. SLAM++~\cite{salas2013slam++}, an RGBD SLAM, uses point pair features to detect and align CAD models into the map. To remove the object template database, several approaches turn to deformable templates~\cite{yingze2013dense, parkhiya2018constructing}.

Learning a shape prior that takes advantage of object shape regularity is another research trend for object shape reconstruction. Intra-class full 3D shape variance is captured in a learned latent space. Object shape is optimised in this latent space given image or depth evidence, and thus full 3D objects can be reconstructed even if only partial observations are available. 
Shape latent space is often learned via (Kernal) PCA~\cite{dambreville2008robust,wang2019directshape} or GP-LVM~\cite{prisacariu2012simultaneous,dame2013dense}. 

Motivated by the success of deep learning in scene recognition, deep networks are used as function approximators that map an image~\cite{fan2017point, tulsiani2017multi} or images~\cite{choy20163d, xie2019pix2vox} to a 3D object shape. Instead of a direct mapping, another line of works~\cite{wu2016learning,zhu2018object,li2019optimizable,lin2019photometric} apply deep networks as powerful dimension compression tools to learn a shape embedding. Object shapes can be optimised in the embedding given visual observations. However, these methods are often constrained to single-object scenes where all observations can be assigned to the same object. When there are multiple objects in a scene (\eg a dining room with a table surrounded by multiple chairs), data association that assigns observations to different objects is essential to apply those methods.

Leveraging a ray clustering based approach for data association, FroDO~\cite{runz2020frodo} demonstrates multi-object reconstruction from monocular image sequences. Although FroDO and the proposed system share common ground on following coarse-to-fine reconstruction, where objects are firstly localised and represented coarsely using cubes/ellipsoids, followed by a dense shape reconstruction, the ray clustering algorithm of FroDO assumes a static environment. In contrast, the proposed system can work with both dynamic and static objects. 
Additionally, \methodname is an on-line approach, whereas FroDO is off-line. 

A number of RGBD based approaches leverage modern instance segmentation networks to fuse depth maps of each object instance separately to achieve object-centric mapping. Assuming a static environment, Fusion++~\cite{mccormac2018fusion++} generates a TSDF reconstruction for each object detected given a RGBD image sequence. MID-Fusion~\cite{xu2019mid} takes a step forward by tracking the pose of each object to handle dynamic objects. Co-Fusion~\cite{runz2017co} and MaskFusion~\cite{runz2018maskfusion} using surfels to represent object shapes can also handle dynamic objects by tracking the object motion using Iterative Closest Point (ICP). Recently, Sucar \etal~\cite{sucar2020neural} propose to optimise object shapes in a learned embedding given multiple depth observations.
In contrast to methods aforementioned focusing on indoor environments, DynSLAM~\cite{barsan2018robust} is a stereo-based system for dynamic object tracking reconstruction in outdoor environments. 
The distinctive advantage of \methodname over these methods is our simple sensory input being a monocular RGB camera.
While \cite{luiten2020track} explores the possibility to replace the depth sensor with a monocular depth estimation network, fusing multiple noisy depth prediction is error-prone. 
An additional benefit of our reconstruction using shape prior is the completeness of the reconstruction.

%% file: c2_method.tex
\subsection{\fixtypo{Notations and Preliminaries}}
In the rest of the paper, we will use the following notation: lower-case bold $\mathbf{t}$ and upper-case bold $\mathbf{T}$ denote a vector and matrix respectively. $\mathbf{T}_{ab}$ denotes the transformation matrix from coordinate frame $b$ to $a$. A vector in coordinate frame $w$ is denoted as $\mathbf{x}^w$.


\subsection{System overview}
Given a new RGB frame, \methodname first employs a monocular 3D detector to predict a 9-DoF object pose, object class label, and 2D bounding box (\ref{subsec:detection}). For each detected object, an image patch cropped by the 2D bounding box of an object is mapped to a shape code in a learned shape embedding (\ref{subsec:shape}). 
State (\ie pose and motion status) of each existing object in the map is modelled by a multiple model Bayesian filter. Prior to data association, we use the filter to predict object location and decide whether an object is matchable using the predicted motion status. 
The new detections are associated with the matchable objects based on a simple but practical pairwise cost (\ie 3D Generalised IoU~\cite{rezatofighi2019generalized}) as the matching cost. 
We solve the linear assignment problem using the Munkres algorithm~\cite{munkres1957algorithms} to decide whether a detection merges to an object track or instantiates a new object in the map. Filters are updated using the associated detections (\ref{subsec:tracking}). 
To reconstruct an object shape, multiple single-view shape codes are fused into a single one by taking the mean, which is decoded by the shape decode to a TSDF. 
The object shape is transformed to the world coordinate using the updated object pose (\ref{subsec:reconstruct}). 
Fig. \ref{fig:pipeline} illustrates the pipeline of our system.
 
\subsection{3D localisation}\label{subsec:detection}
First, \methodname detects objects of interest given an RGB image. We apply a monocular 3D detector that takes a single RGB image as input and outputs both 2D attributes (\ie object class and 2D bounding box) and 3D attributes (\ie object translation $\mathbf{t}_{co}$ and viewpoint $\mathbf{R}_{co}$ with respect to the camera, and object 3D dimension $(s_x, s_y, s_z)$).
Technically, the detector is trained to predict an offset $(\Delta x, \Delta y)$ between the center of the 2D bounding box $(x_{2d}, y_{2d})$ and the projection center of the 3D shape $(x_{3d}, y_{3d})$ on the image plane. We also predict the object depth value $z$. Assuming we know the camera intrinsic parameters $f_x$, $f_y$, $c_x$, $c_y$, the object's 3D center $\mathbf{t}_{co}$ in camera coordinate frame is recovered as follows:
\begin{align}
    \mathbf{t}_{co}^T &= [\frac{x_{2d} + \Delta x - c_x}{f_x} z, \frac{y_{2d} + \Delta y - c_y}{f_y} z, z]
\end{align}

To handle the multi-modal nature of symmetric objects, we reformulate object viewpoint prediction as a classification problem, where azimuth $\mathbf{R}_{azi}$ and elevation $\mathbf{R}_{ele}$ are discretised into $36$ and $10$ bins respectively.
The rotation matrix is $\mathbf{R}_{co} = \mathbf{R}_{ele} \mathbf{R}_{azi}$.
The transformation matrix $\mathbf{T}_{co} \in \mathrm{SE(3)}$ from the canonical object space to camera coordinate frame is:
\begin{align}
    \mathbf{T}_{co} = \begin{bmatrix}
    \mathbf{R}_{co} & \mathbf{t}_{co} \\
    \mathbf{0} & 1
    \end{bmatrix}
\end{align}
Together with the scale parameters, each detected object is localised as an oriented 3D bounding in the camera coordinate frame.

\subsection{Shape embedding and inference}\label{subsec:shape}
As a shape prior based reconstruction, we are interested in reconstructing the complete object shape even if only a partial observation is available. The formulation of our shape embedding and inference follows FroDO~\cite{runz2020frodo} closely. 
We use compact k-dimensional shape codes $\mathbf{l} \in \mathbb{R}^{k}$ embedded in a learned latent space to parameterise normalised object shapes in a canonical pose throughout our system. This latent representation effectively allows us to leverage the learned latent space as a shape prior. A TSDF, where the zero-crossing level set is the object surface, can be decoded from the latent code via a DeepSDF decoder $G(\mathbf{l})$~\cite{park2019deepsdf}.

After each object has been detected, we estimate a single-view shape code by mapping its cropped 2D bounding box to the shape embedding using the shape encoder. 
Note that at this point we do not reconstruct the shape by decoding the single-view shape code; instead, the shape is decoded later, once shape codes have been fused over time, as described in \ref{subsec:reconstruct}.

\subsection{Tracking}\label{subsec:tracking}
Because single-view detections are mostly noisy, a common approach in MOT is to apply a Bayesian filter on the object motion to smooth the tracking trajectory~\cite{weng2020ab3dmot, shenoi2020jrmot}, and to provide motion predictions for better data association.

To deal with both dynamic and static objects, because there is no one-size-fit-all motion model for use within a Bayesian filter, we employ the well-known Interacting Multiple Models (IMM) filter~\cite{blom1988interacting}, in which we maintain kinematics and a motion status selection variable under the assumption of linear kinematic and observation model and Gaussian noise.
We can further leverage the motion status to finesse thresholds associated with the persistence of a trajectory when there is no observation 

Similar to any MOT approach that has to handle trajectory birth and death, we deal with trajectory birth via the standard method, which is instantiating a tentative trajectory from a detection. A tentative trajectory is upgraded to a confirmed trajectory only if it is observed $n$ consecutive times. We treat trajectory termination differently. The death of an object trajectory is not controlled by a predefined fixed time threshold. Instead, an object trajectory is terminated only if it is a dynamic object and it is not observed in the last $n$ frames (\ie static objects remain in the map even if not observed).
\revise{If an object instance is moved during an unobserved period, it is considered a new object instance when re-observed. The old object instance (if the old position is visible later) is pruned using negative information. Technically, if the largest 2D IoU between the projection of a 3D object shape and all detected 2D bounding boxes of an image is less than $0.5$, this object is considered not visible and pruned for this time step.}

For each RGB frame, while the 3D detector returns a set of new detections, the filter predicts each existing object's location and motion status at the current frame. The next step is to associate the new detections to the existing objects.
We construct a $M \times N$ cost matrix between $M$ new detections in the current frame and $N$ matchable objects in the map, where each element represents the cost for associating the $m^{th}$ detected objects to the $n^{th}$ objects in the map, and the cost is measured by the negative of pairwise 3D GIoU~\cite{rezatofighi2019generalized}.
To calculate the GIoU, the detections are transformed to world coordinate where the existing objects are. The optimal matching is found by the Munkres algorithm~\cite{munkres1957algorithms} with gating (\ie A pair of detection and object is considered matched if the cost is below a fixed threshold). 
The filters are updated using the associated detections. Details of IMM filter prediction and update can be referred to \cite{blom1988interacting}.

\subsection{Reconstruction}\label{subsec:reconstruct}
After object tracking, the last step of \methodname at each RGB frame is to reconstruct each object's dense shape in the map. We fuse all single-view shape codes up to the current frame by averaging them into a single code $\mathbf{l}_f = \frac{1}{N}\sum_{i}^N \mathbf{l}_i$. 
A TSDF that represents an object shape in the canonical object coordinate is decoded from the shape decoder $\mathbf{X}^o = G(\mathbf{l}_f)$ given the fused shape code. 
An object shape mesh is extracted from the TSDF by the Marching Cube algorithm~\cite{lorensen1987marching}. 
The mesh is then transformed to the world coordinate using updated pose from object tracking:
\begin{align}
    \mathbf{X}^w&=\mathbf{T}_{wo} \mathbf{S} \mathbf{X}^o \\
    \mathbf{S} &= \begin{bmatrix}
        s_x & 0 & 0 \\
        0 & s_y & 0 \\
        0 & 0 & s_z \\
        \end{bmatrix}
\end{align}
where $\mathbf{T}_{wo}$ is the rigid transformation from object coordinate to world coordinate and $\mathbf{S}$ is the scale matrix.

The shape and pose can be further optimised using visual cues, such as silhouette and photometric consistency, as done in FroDO~\cite{runz2020frodo}, but the optimisation is out of the scope of this paper.



\begin{figure*}[th]
    \centering
    \includegraphics[width=0.99\linewidth]{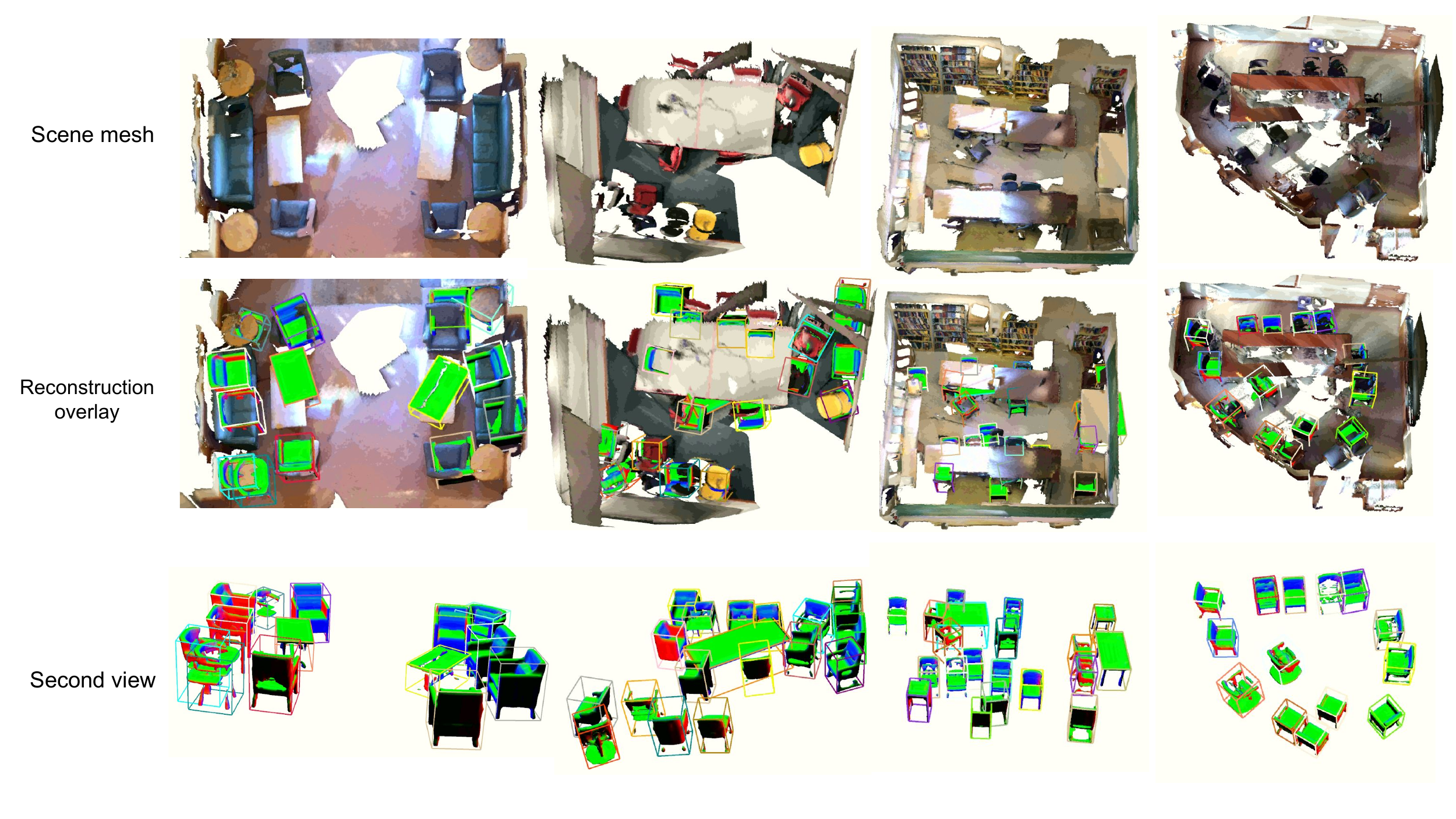}
    \caption{\small{Localisation and reconstruction on ScanNet~\cite{dai2017scannet} sequences. Top row: Ground-truth scan mesh for reference, middle row: objects overlay on the ground-truth mesh to show localisation quality. Bottom row: object shape reconstruction. Scan mesh is for visualization purpose only. The input to \methodname is camera poses and RGB images only. }}
    \label{fig:scannet_reconstruct}
    \vspace{-0.3cm}
\end{figure*}

\subsection{Implementation Details}
We dedicate this section to describe the implementation details of each component in the pipeline.

Our detector is built on top of a 2D detector DETR~\cite{carion2020end}. To predict the additional 3D attributes, we extend the original DETR by adding an independent prediction feed-forward network for each additional attribute.  
For indoor scenes, we train the 3D detector on the ScanNet~\cite{dai2017scannet} images. Because ScanNet annotations do not provide object dimensions and orientation that we need for training our detector, we use the CAD model annotation provided by Scan2CAD~\cite{Avetisyan2019scan2cad}. We finetune the detector from the official release on the official ScanNet train/val split for $10$ epochs.
For outdoor scenes, we use an off-the-shelf detector from ~\cite{zhou2020tracking}.

We use $k=64$ dimensions for our shape embedding. The architecture of our shape decoder is identical to the original DeepSDF~\cite{park2019deepsdf}, and we closely follow DeepSDF in the training procedure. The only difference is that we train separate embeddings for indoor (\eg chair, table and display), and outdoor scenes (\eg car).
The architecture of our shape encoder is modified from the ResNet18 by changing the original output dimension to our embedding dimension.
It is trained on synthetic images rendered from the ShapeNet~\cite{chang2015shapenet} CAD models with random backgrounds.

We formulate the state vector of the IMM filter as a 7-dimensional vector $(\mathbf{\mu}, \pi)$, where $\mathbf{\mu}=[c_x, c_y, c_z, v_x, v_y, v_z]$ is a 6-dimension vector that represents the centre and velocity of an object and $\pi$ is the model selection variable. 
Object center observation is from the monocular 3D detector, and the measurement covariance is set to $0.01 \mathbf{I} \in \mathbb{R}^{3\times3}$ and $0.25 \mathbf{I} \in \mathbb{R}^{3\times3}$ for indoor and outdoor environment respectively. 
We use zero velocity and the first observation to initialise the state mean, and covariance is initialised using identity matrix $\mathbf{I} \in \mathbb{R}^{6\times6}$. 
We use a constant velocity model (with acceleration as process noise) and zero velocity model (also known as random walk) for dynamic and static motion model respectively. 
The transition probability matrix is set to$[[0.6, 0.4];[0.4, 0.6]]$. 
An object is classified as static if $p(\mathit{\pi}=static) > p(\mathit{\pi}=dynamic)$.
Note that at present, we do not incorporate object rotation in the filter state. 
One complexity of doing so is that the noise of rotation observation predicted by the deep network is non-Gaussian due to object symmetry. 
For instance, given a car's side-view image, the network prediction has two modes being the left and right side of the car. 
We attempt to capture this categorical distribution using a Particle Filter for object rotation, but it is discarded later in the development due to the concern about speed.

The data association gating threshold measured by the 3D Generalised IoU is set to $1.75$ for outdoor scenes and $0.25$ for indoor scenes. We need a higher threshold in the outdoor environments to accommodate that outdoor objects move faster.

We use ground-truth camera poses in our experiments on KITTI and ScanNet for a fair comparison, but the proposed system can work with off-the-shelf SLAM systems to obtain camera poses. 
To this end, we demonstrate that our system can work with estimated camera trajectory from DF-VO~\cite{zhan2020visual} on KITTI dataset.

%% file: c3_experiment.tex

\subsection{Datasets}
We quantitatively evaluate \methodname on KITTI~\cite{Geiger2012KITTI} and ScanNet~\cite{dai2017scannet}. KITTI is a popular dataset used for object tracking benchmark in outdoor scenes. It consists of image sequences captured by a camera mounted on a moving vehicle in different road conditions. 
In contrast, sequences in ScanNet are captured in various indoor scenes (\eg offices, living rooms, or conference rooms) using a handheld device. 
However, because the annotated bounding boxes in ScanNet annotation is subject to occlusions or reconstruction failures that lead to incomplete bounding boxes, following FroDO, we use the annotations from Scan2CAD~\cite{Avetisyan2019scan2cad} to obtain amodal 3D bounding boxes for evaluation. 
Since objects in ScanNet are static, we demonstrate indoor dynamic objects using self-recorded videos qualitatively.

\begin{figure*}[t]
    \centering
    \includegraphics[width=0.95\linewidth]{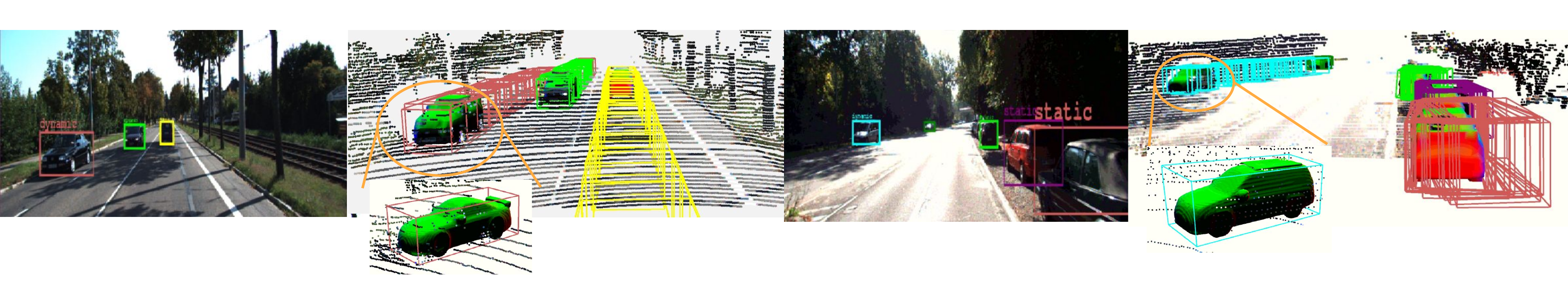}
    \caption{\small{Object tracking on KITTI dataset~\cite{Geiger2012KITTI}. The tracking is consistent, and we correctly label dynamic/static objects using the mode selection variable in the IMM filter. The reconstruction of vehicles are also highlighted. Lidar points are used for visualization purpose.}}
    \label{fig:kitti_tracking}
    \vspace{-0.3cm}
\end{figure*}

\subsection{Localisation}
We compare \methodname with FroDO~\cite{runz2020frodo} on object localisation in 3D to demonstrate the effect of our monocular 3D detector and data association.  
Table \ref{tab:detection} presents the comparison with FroDO on three common object categories (\ie, chair, table and display) in indoor scenes. 
The evaluation metric is the widely adopted mean Average Precision (mAP) in object detection, and the Intersection over Union threshold is set to 0.5.

We outperform FroDO on both chair and display class and have similar performance on table class. 
We believe that the improvement is due to the differences in our detection and data association approach. 
FroDO used 2D detections and a ray clustering approach for data association. 
The 3D bounding boxes are obtained by triangulating associated 2D detections. 
The ray clustering based data association suffers from local minima and leads to incorrect matching if objects are close to each other. 
\methodname circumvents this problem by using monocular 3D detection, and thus the following data association works in the 3D space directly. 
Qualitative results of \methodname on ScanNet is shown in Fig. \ref{fig:scannet_reconstruct}. 
It can be seen that the proposed method is more effective on chairs than tables (consistent with the results shown in Table~\ref{tab:detection}), large dining table or conference table in particular.
This is because the 3D detector tends to perform worse on truncated views (that cannot observe the full extent of an object) of objects. 
If an image only captures a corner of a table, it is hard to determine how far this table would extend beyond the image frame. 
Hence, the predictions on dimension and object translation are more uncertain.

\begin{table}[t]
  \begin{center}
  \small
    \begin{tabular}{l|c|c|c} 
    \hline
      mAP @ IoU=0.5 & Chair $\uparrow$ & Table $\uparrow$ & Display $\uparrow$ \\
      \hline\hline
      FroDO \cite{runz2020frodo} & $0.32$ & $0.06$ & $0.04$\\
      Ours & $\mathbf{0.39}$ & $0.06$ & $\mathbf{0.10}$ \\
      \hline
    \end{tabular}
    \caption{\small{3D detection comparison on \fixtypo{ScanNet~\cite{dai2017scannet}}}}
    \label{tab:detection}
  \end{center}
  \vspace{-0.5cm}
\end{table}

\subsection{Tracking}
In the conventional KITTI benchmark evaluation, results of 3D MOT are evaluated following the 2D MOT evaluation, where 3D tracking results are projected to the image plane for evaluation. Therefore, it fails to reflect errors on depth direction (\ie an object located at any point on the projection ray results in the same error). We instead use the 3D MOT evaluation recently proposed in AB3DMOT~\cite{weng2020ab3dmot}. The evaluation metrics are Multiple Object Tracking Accuracy (MOTA), Multiple Object Tracking Precision (MOTP) and ID Switches (IDS). 

We use CenterTrack~\cite{zhou2020tracking}, a monocular 3D multiple object tracking framework, as a baseline. While we share the same monocular 3D detector, the main difference is in the motion tracking and data association method. CenterTrack predicts the 2D object motion on the image plane using a deep network, and associates detections between adjacent frames using the IoU between 2D bounding boxes as a matching cost. Matches are found by a greedy search. We model each object motion using an independent IMM filter in the 3D space, and choose Munkres algorithm~\cite{munkres1957algorithms} over a greedy search.

The quantitative result is shown in Table \ref{tab:tracking}.
We believe the reason for the improvement \fixtypo{is twofold}: 1) We perform object tracking and data association in the 3D space so objects in similar depth direction but with different values can be separated easily; 2) we solve the linear assignment problem in data association instead of a greedy search. 

\revise{We study the effect of measurement noise. Increasing the measurement noise leads to over-smoothing for the object trajectory and deteriorates performance. Although lowering the measurement noise does not affect MOTA and MOTP significantly, it increases the IDS due to the noise in single-view 3D detection.}
The object ID consistency is crucial for object-centric mapping as duplicate object instances degenerate the reconstruction quality. We visualise our tracking result and the motion state estimation by the IMM filter in Fig. \ref{fig:kitti_tracking}. More qualitative results of object tracking, motion state estimation and reconstruction on KITTI dataset are shown in the supplementary video.

To verify that our approach can also work with estimated camera poses from a SLAM or VO system, we also run evaluation with estimated camera poses from the state-of-the-art Visual Odometry system -- DF-VO~\cite{zhan2020visual}. 
The performance of both CenterTrack and our method drops slightly due to the camera tracking error, but note that ours still outperform CenterTrack.

\begin{table}[t]
  \begin{center}
  \small
    \begin{tabular}{l|c|c|c} 
    \hline
      Methods & MOTA $\uparrow$ & MOTP $\uparrow$ & IDS $\downarrow$ \\
      \hline\hline
      CenterTrack~\cite{zhou2020tracking} & $0.34$ & $0.53$ & $68$ \\
      Ours (No filter) & $0.37$ & $0.53$ & $12$ \\
      \revise{Ours ($\sigma$=0.1)} & \revise{$0.39$} & \revise{$0.53$} & \revise{7} \\
      \revise{Ours ($\sigma$=1)} & \revise{0.36} & \revise{0.51} & \revise{4} \\
      Ours \revise{($\sigma$=0.25)} & $\mathbf{0.39}$ & $0.53$ & $\mathbf{0}$ \\
      \hline\hline
      \revise{CenterTrack~\cite{zhou2020tracking}(w/VO)} & \revise{$0.31$} & \revise{$0.51$} & \revise{$68$} \\
      \revise{Ours (w/VO)} & \revise{$0.36$} & \revise{$0.52$} & \revise{$0$} \\
      \hline
    \end{tabular}
    \caption{\small{3D Object Tracking comparison on \fixtypo{KITTI~\cite{Geiger2012KITTI}}, \revise{$\sigma$ is multiplier on the diagonal covariance matrix.}}}
    \label{tab:tracking}
  \end{center}
  \vspace{-0.5cm}
\end{table}

\subsection{Reconstruction}
\begin{table*}[t]
  \begin{center}
  \small
    \begin{tabular}{l|c c c c | c c c}
    \hline
    Methods &\multicolumn{4}{| c |}{Error metrics} & \multicolumn{3}{c}{accuracy} \\ 
    & RMSE $\downarrow$ & log RMSE $\downarrow$ & Abs Rel $\downarrow$ & Seq Rel $\downarrow$ & $\delta < 1.25$ $\uparrow$ & $\delta < 1.25^2$ $\uparrow$ & $\delta < 1.25^3$ $\uparrow$ \\
    \hline\hline
    MOTSFusion (Mono.) & $5.33$ & $0.26$ & $0.17$ & $1.71$ & $0.76$ & $0.91$ & $0.94$ \\
    Ours & $\mathbf{2.09}$ & $\mathbf{0.13}$ & $\mathbf{0.12}$ & $\mathbf{0.50}$ & $\mathbf{0.88}$ & $\mathbf{0.97}$ & $\mathbf{0.99}$ \\
    \hline
    \end{tabular}
    \caption{\small{Depth comparison to MOTSFusion on \fixtypo{KITTI~\cite{Geiger2012KITTI}}}}
    \label{tab:depth}
  \end{center}
  \vspace{-0.3cm}
\end{table*}

We compare \methodname against the monocular MOTSFusion, where they use a monocular depth estimation network followed by an instance segmentation network for object reconstruction to recover the \emph{visible} surfaces of objects. Because MOTSFusion does not reconstruct full 3D shape, our method would be favoured if we were to evaluate reconstruction in 3D. Instead for a fairer comparison, we render our full 3D shape onto the image plane as a depth map, and compare the reconstruction quality against MOTSFusion via depth map evaluation.

\begin{figure*}[t]
    \centering
    \includegraphics[width=0.9\linewidth]{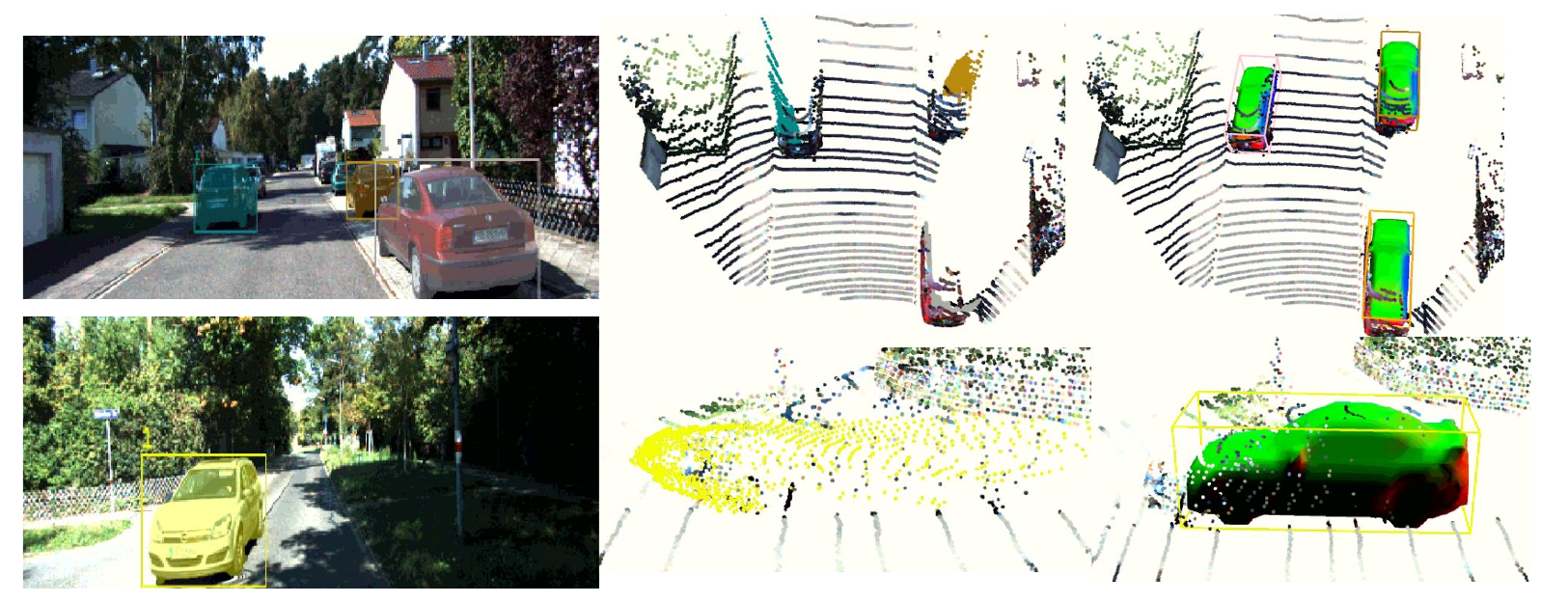}
    \caption{\small{Reconstruction comparison to monocular MOTSFusion on KITTI. Left column: Current frame, middle column: reconstruction by MOTSFusion, right column: our reconstruction. Note that colored lidar points are used for visualization only, not part of the processing.}}
    \label{fig:depth_compare}
    \vspace{-0.3cm}
\end{figure*}

Our shape prior driven approach outperforms MOTSFusion by a large margin, as shown in Table \ref{tab:depth}. MOTSFusion particularly suffers from RMSE, indicating it is affected by the blurry object edge from the depth prediction and instance segmentation. To better contrast both methods, we visualise the comparison in Fig. \ref{fig:depth_compare}. 
Even when MOTSFusion can reconstruct the surface accurately, our shape prior based reconstruction is still advantageous as we can reconstruct the full 3D shape of an object.

\subsection{Indoor tracking and reconstruction}
We run \methodname on a self-recorded video with the focus on demonstrating tracking objects whose motion status switches between dynamic and static throughout the video. \methodname is able to classify whether an object is static or dynamic accurately at each time step, which indicates that our IMM filter captures the model switching behaviour. 
Another highlight of this video is that a chair is tracked successfully even it was completely occluded by another object for a period of time. 
That is because the chair in the red bounding box is classified as static correctly, and we use the motion status to control object trajectory termination. 
The occluded chair is at risk of being discarded if we follow common MOT practice of a predefined fixed time threshold to terminate unobserved objects.

\subsection{Runtime analysis}
All experiments are run on an Intel Core i7 desktop with 16 GB RAM and an Nvidia GeForce GTX 1070 GPU. The runtime cost for each component is shown in Table \ref{tab:runtime}. In practice, we can run between 2 - 4 Hz in a scene with 5 objects.

\begin{table}[t]
  \begin{center}
  \small
    \begin{tabular}{l|c|c|c|c} 
    \hline
      stage & Detection & \makecell{shape \\encode} & association & \makecell{shape \\decode} \\
      \hline\hline
      \makecell{time\\(ms)} & 111/frame & 4/det. obj. & 3/frame & 35/obj.\\
      \hline
    \end{tabular}
    \caption{\small{Runtime analysis breakdown for each system component. det. obj. refers to detected object}}
    \label{tab:runtime}
  \end{center}
  \vspace{-0.5cm}
\end{table}

%% file: c4_conclusion.tex
In this paper, we presented \methodname, a framework for multi-object localization, tracking and reconstruction given monocular image sequences. 
We leverage the deep shape prior for complete and accurate shape reconstruction and the IMM filter to jointly track the motion of an object and discriminate motion status. 
We evaluate \methodname extensively on both indoor and outdoor scenes under both static and dynamic environment. 
While we show that the data association, which relies on the 3D GIoU, is practical, an interesting future direction is to develop a learning-based approach for data association. 
This could furthermore pave the way for an end-to-end learnable system.
\revise{We sometimes observe that the filter over-smooths object motion due to a predefined high measurement noise to handle the noisy single-view detections. It is worth exploring probabilistic object detection to improve the tracking performance.}
\methodname has benefited from SLAM to provide camera poses. Another promising future direction is to integrate \methodname into a SLAM framework such that the object prior knowledge could be leveraged in SLAM.

%% file: RA-L-Template-YY.bbl
\begin{thebibliography}{10}
\providecommand{\url}[1]{#1}
\csname url@rmstyle\endcsname
\providecommand{\newblock}{\relax}
\providecommand{\bibinfo}[2]{#2}
\providecommand\BIBentrySTDinterwordspacing{\spaceskip=0pt\relax}
\providecommand\BIBentryALTinterwordstretchfactor{4}
\providecommand\BIBentryALTinterwordspacing{\spaceskip=\fontdimen2\font plus
\BIBentryALTinterwordstretchfactor\fontdimen3\font minus
  \fontdimen4\font\relax}
\providecommand\BIBforeignlanguage[2]{{%
\expandafter\ifx\csname l@#1\endcsname\relax
\typeout{** WARNING: IEEEtran.bst: No hyphenation pattern has been}%
\typeout{** loaded for the language `#1'. Using the pattern for}%
\typeout{** the default language instead.}%
\else
\language=\csname l@#1\endcsname
\fi
#2}}

\bibitem{klein2007parallel}
G.~Klein and D.~Murray, ``Parallel tracking and mapping for small ar
  workspaces,'' in \emph{2007 6th IEEE and ACM international symposium on mixed
  and augmented reality}.\hskip 1em plus 0.5em minus 0.4em\relax IEEE, 2007,
  pp. 225--234.

\bibitem{davison2007monoslam}
A.~J. Davison, I.~D. Reid, N.~D. Molton, and O.~Stasse, ``{MonoSLAM}: Real-time
  single camera slam,'' \emph{IEEE transactions on pattern analysis and machine
  intelligence}, vol.~29, no.~6, pp. 1052--1067, 2007.

\bibitem{newcombe2011dtam}
R.~A. Newcombe, S.~J. Lovegrove, and A.~J. Davison, ``{DTAM}: Dense tracking
  and mapping in real-time,'' in \emph{2011 international conference on
  computer vision}.\hskip 1em plus 0.5em minus 0.4em\relax IEEE, 2011, pp.
  2320--2327.

\bibitem{newcombe2011kinectfusion}
R.~A. Newcombe, S.~Izadi, O.~Hilliges, D.~Molyneaux, D.~Kim, A.~J. Davison,
  P.~Kohi, J.~Shotton, S.~Hodges, and A.~Fitzgibbon, ``{KinectFusion}:
  Real-time dense surface mapping and tracking,'' in \emph{2011 10th IEEE
  International Symposium on Mixed and Augmented Reality}.\hskip 1em plus 0.5em
  minus 0.4em\relax IEEE, 2011, pp. 127--136.

\bibitem{whelan2015elasticfusion}
T.~Whelan, S.~Leutenegger, R.~Salas-Moreno, B.~Glocker, and A.~Davison,
  ``{ElasticFusion}: Dense slam without a pose graph.''\hskip 1em plus 0.5em
  minus 0.4em\relax Robotics: Science and Systems.

\bibitem{stuckler2014multi}
J.~St{\"u}ckler and S.~Behnke, ``Multi-resolution surfel maps for efficient
  dense 3d modeling and tracking,'' \emph{Journal of Visual Communication and
  Image Representation}, vol.~25, no.~1, pp. 137--147, 2014.

\bibitem{hermans2014dense}
A.~Hermans, G.~Floros, and B.~Leibe, ``Dense 3d semantic mapping of indoor
  scenes from rgb-d images,'' in \emph{2014 IEEE International Conference on
  Robotics and Automation (ICRA)}.\hskip 1em plus 0.5em minus 0.4em\relax IEEE,
  2014, pp. 2631--2638.

\bibitem{pham2015hierarchical}
T.~T. Pham, I.~Reid, Y.~Latif, and S.~Gould, ``Hierarchical higher-order
  regression forest fields: An application to 3d indoor scene labelling,'' in
  \emph{Proceedings of the IEEE international conference on computer vision},
  2015, pp. 2246--2254.

\bibitem{mccormac2017semanticfusion}
J.~McCormac, A.~Handa, A.~Davison, and S.~Leutenegger, ``Semanticfusion: Dense
  3d semantic mapping with convolutional neural networks,'' in \emph{2017 IEEE
  International Conference on Robotics and automation (ICRA)}.\hskip 1em plus
  0.5em minus 0.4em\relax IEEE, 2017, pp. 4628--4635.

\bibitem{galvez2016real}
D.~G{\'a}lvez-L{\'o}pez, M.~Salas, J.~D. Tard{\'o}s, and J.~Montiel,
  ``Real-time monocular object slam,'' \emph{Robotics and Autonomous Systems},
  vol.~75, pp. 435--449, 2016.

\bibitem{salas2013slam++}
R.~F. Salas-Moreno, R.~A. Newcombe, H.~Strasdat, P.~H. Kelly, and A.~J.
  Davison, ``Slam++: Simultaneous localisation and mapping at the level of
  objects,'' in \emph{Proceedings of the IEEE conference on computer vision and
  pattern recognition}, 2013, pp. 1352--1359.

\bibitem{yingze2013dense}
S.~Yingze~Bao, M.~Chandraker, Y.~Lin, and S.~Savarese, ``Dense object
  reconstruction with semantic priors,'' in \emph{Proceedings of the IEEE
  Conference on Computer Vision and Pattern Recognition}, 2013, pp. 1264--1271.

\bibitem{parkhiya2018constructing}
P.~Parkhiya, R.~Khawad, J.~K. Murthy, B.~Bhowmick, and K.~M. Krishna,
  ``{Constructing category-specific models for monocular object-SLAM},'' in
  \emph{2018 IEEE International Conference on Robotics and Automation
  (ICRA)}.\hskip 1em plus 0.5em minus 0.4em\relax IEEE, 2018, pp. 1--9.

\bibitem{dambreville2008robust}
S.~Dambreville, R.~Sandhu, A.~Yezzi, and A.~Tannenbaum, ``Robust 3d pose
  estimation and efficient 2d region-based segmentation from a 3d shape
  prior,'' in \emph{European Conference on Computer Vision}.\hskip 1em plus
  0.5em minus 0.4em\relax Springer, 2008, pp. 169--182.

\bibitem{wang2019directshape}
R.~Wang, N.~Yang, J.~Stueckler, and D.~Cremers, ``Directshape: Photometric
  alignment of shape priors for visual vehicle pose and shape estimation,''
  \emph{arxiv}, pp. arXiv--1904, 2019.

\bibitem{prisacariu2012simultaneous}
V.~A. Prisacariu, A.~V. Segal, and I.~Reid, ``Simultaneous monocular 2d
  segmentation, 3d pose recovery and 3d reconstruction,'' in \emph{Asian
  conference on computer vision}.\hskip 1em plus 0.5em minus 0.4em\relax
  Springer, 2012, pp. 593--606.

\bibitem{dame2013dense}
A.~Dame, V.~A. Prisacariu, C.~Y. Ren, and I.~Reid, ``Dense reconstruction using
  3d object shape priors,'' in \emph{Proceedings of the IEEE Conference on
  Computer Vision and Pattern Recognition}, 2013, pp. 1288--1295.

\bibitem{fan2017point}
H.~Fan, H.~Su, and L.~J. Guibas, ``A point set generation network for 3d object
  reconstruction from a single image,'' in \emph{Proceedings of the IEEE
  conference on computer vision and pattern recognition}, 2017, pp. 605--613.

\bibitem{tulsiani2017multi}
S.~Tulsiani, T.~Zhou, A.~A. Efros, and J.~Malik, ``Multi-view supervision for
  single-view reconstruction via differentiable ray consistency,'' in
  \emph{Proceedings of the IEEE conference on computer vision and pattern
  recognition}, 2017, pp. 2626--2634.

\bibitem{choy20163d}
C.~B. Choy, D.~Xu, J.~Gwak, K.~Chen, and S.~Savarese, ``3d-r2n2: A unified
  approach for single and multi-view 3d object reconstruction,'' in
  \emph{European conference on computer vision}.\hskip 1em plus 0.5em minus
  0.4em\relax Springer, 2016, pp. 628--644.

\bibitem{xie2019pix2vox}
H.~Xie, H.~Yao, X.~Sun, S.~Zhou, and S.~Zhang, ``Pix2vox: Context-aware 3d
  reconstruction from single and multi-view images,'' in \emph{Proceedings of
  the IEEE International Conference on Computer Vision}, 2019, pp. 2690--2698.

\bibitem{wu2016learning}
J.~Wu, C.~Zhang, T.~Xue, B.~Freeman, and J.~Tenenbaum, ``Learning a
  probabilistic latent space of object shapes via 3d generative-adversarial
  modeling,'' in \emph{Advances in neural information processing systems},
  2016, pp. 82--90.

\bibitem{zhu2018object}
R.~Zhu, C.~Wang, C.-H. Lin, Z.~Wang, and S.~Lucey, ``Object-centric photometric
  bundle adjustment with deep shape prior,'' in \emph{2018 IEEE Winter
  Conference on Applications of Computer Vision (WACV)}.\hskip 1em plus 0.5em
  minus 0.4em\relax IEEE, 2018, pp. 894--902.

\bibitem{li2019optimizable}
K.~Li, R.~Garg, M.~Cai, and I.~Reid, ``Single-view object shape reconstruction
  using deep shape prior and silhouette,'' in \emph{30th British Machine Vision
  Conference 2019, Cardiff, UK}.\hskip 1em plus 0.5em minus 0.4em\relax {BMVA}
  Press, 2019, p. 163.

\bibitem{lin2019photometric}
C.-H. Lin, O.~Wang, B.~C. Russell, E.~Shechtman, V.~G. Kim, M.~Fisher, and
  S.~Lucey, ``Photometric mesh optimization for video-aligned 3d object
  reconstruction,'' in \emph{Proceedings of the IEEE Conference on Computer
  Vision and Pattern Recognition}, 2019, pp. 969--978.

\bibitem{runz2020frodo}
M.~Runz, K.~Li, M.~Tang, L.~Ma, C.~Kong, T.~Schmidt, I.~Reid, L.~Agapito,
  J.~Straub, S.~Lovegrove, \emph{et~al.}, ``{FroDO}: From detections to 3d
  objects,'' in \emph{Proceedings of the IEEE/CVF Conference on Computer Vision
  and Pattern Recognition}, 2020, pp. 14\,720--14\,729.

\bibitem{mccormac2018fusion++}
J.~McCormac, R.~Clark, M.~Bloesch, A.~Davison, and S.~Leutenegger, ``Fusion++:
  Volumetric object-level slam,'' in \emph{2018 international conference on 3D
  vision (3DV)}.\hskip 1em plus 0.5em minus 0.4em\relax IEEE, 2018, pp. 32--41.

\bibitem{xu2019mid}
B.~Xu, W.~Li, D.~Tzoumanikas, M.~Bloesch, A.~Davison, and S.~Leutenegger,
  ``{MID-Fusion}: Octree-based object-level multi-instance dynamic slam,'' in
  \emph{2019 International Conference on Robotics and Automation (ICRA)}.\hskip
  1em plus 0.5em minus 0.4em\relax IEEE, 2019, pp. 5231--5237.

\bibitem{runz2017co}
M.~R{\"u}nz and L.~Agapito, ``{Co-fusion}: Real-time segmentation, tracking and
  fusion of multiple objects,'' in \emph{2017 IEEE International Conference on
  Robotics and Automation (ICRA)}.\hskip 1em plus 0.5em minus 0.4em\relax IEEE,
  2017, pp. 4471--4478.

\bibitem{runz2018maskfusion}
M.~Runz, M.~Buffier, and L.~Agapito, ``Maskfusion: Real-time recognition,
  tracking and reconstruction of multiple moving objects,'' in \emph{2018 IEEE
  International Symposium on Mixed and Augmented Reality (ISMAR)}.\hskip 1em
  plus 0.5em minus 0.4em\relax IEEE, 2018, pp. 10--20.

\bibitem{sucar2020neural}
E.~Sucar, K.~Wada, and A.~Davison, ``Neural object descriptors for multi-view
  shape reconstruction,'' \emph{arXiv preprint arXiv:2004.04485}, 2020.

\bibitem{barsan2018robust}
I.~A. B{\^a}rsan, P.~Liu, M.~Pollefeys, and A.~Geiger, ``Robust dense mapping
  for large-scale dynamic environments,'' in \emph{2018 IEEE International
  Conference on Robotics and Automation (ICRA)}.\hskip 1em plus 0.5em minus
  0.4em\relax IEEE, 2018, pp. 7510--7517.

\bibitem{luiten2020track}
J.~Luiten, T.~Fischer, and B.~Leibe, ``Track to reconstruct and reconstruct to
  track,'' \emph{IEEE Robotics and Automation Letters}, vol.~5, no.~2, pp.
  1803--1810, 2020.

\bibitem{rezatofighi2019generalized}
H.~Rezatofighi, N.~Tsoi, J.~Gwak, A.~Sadeghian, I.~Reid, and S.~Savarese,
  ``Generalized intersection over union: A metric and a loss for bounding box
  regression,'' in \emph{Proceedings of the IEEE Conference on Computer Vision
  and Pattern Recognition}, 2019, pp. 658--666.

\bibitem{munkres1957algorithms}
J.~Munkres, ``Algorithms for the assignment and transportation problems,''
  \emph{Journal of the society for industrial and applied mathematics}, vol.~5,
  no.~1, pp. 32--38, 1957.

\bibitem{park2019deepsdf}
J.~J. Park, P.~Florence, J.~Straub, R.~Newcombe, and S.~Lovegrove, ``{DeepSDF}:
  Learning continuous signed distance functions for shape representation,'' in
  \emph{Proceedings of the IEEE Conference on Computer Vision and Pattern
  Recognition}, 2019, pp. 165--174.

\bibitem{weng2020ab3dmot}
X.~Weng, J.~Wang, D.~Held, and K.~Kitani, ``{AB3DMOT}: A baseline for 3d
  multi-object tracking and new evaluation metrics,'' \emph{arXiv preprint
  arXiv:2008.08063}, 2020.

\bibitem{shenoi2020jrmot}
A.~Shenoi, M.~Patel, J.~Gwak, P.~Goebel, A.~Sadeghian, H.~Rezatofighi,
  R.~Martin-Martin, and S.~Savarese, ``{JRMOT}: A real-time 3d multi-object
  tracker and a new large-scale dataset,'' \emph{arXiv preprint
  arXiv:2002.08397}, 2020.

\bibitem{blom1988interacting}
H.~A. Blom and Y.~Bar-Shalom, ``The interacting multiple model algorithm for
  systems with markovian switching coefficients,'' \emph{IEEE transactions on
  Automatic Control}, vol.~33, no.~8, pp. 780--783, 1988.

\bibitem{lorensen1987marching}
W.~E. Lorensen and H.~E. Cline, ``Marching cubes: A high resolution 3d surface
  construction algorithm,'' \emph{ACM siggraph computer graphics}, vol.~21,
  no.~4, pp. 163--169, 1987.

\bibitem{dai2017scannet}
A.~Dai, A.~X. Chang, M.~Savva, M.~Halber, T.~Funkhouser, and M.~Nie{\ss}ner,
  ``{ScanNet}: Richly-annotated 3d reconstructions of indoor scenes,'' in
  \emph{Proc. Computer Vision and Pattern Recognition (CVPR), IEEE}, 2017.

\bibitem{carion2020end}
N.~Carion, F.~Massa, G.~Synnaeve, N.~Usunier, A.~Kirillov, and S.~Zagoruyko,
  ``End-to-end object detection with transformers,'' \emph{arXiv preprint
  arXiv:2005.12872}, 2020.

\bibitem{Avetisyan2019scan2cad}
A.~Avetisyan, M.~Dahnert, A.~Dai, M.~Savva, A.~X. Chang, and M.~Niessner,
  ``{Scan2CAD}: Learning cad model alignment in rgb-d scans,'' in \emph{The
  IEEE Conference on Computer Vision and Pattern Recognition (CVPR)}, June
  2019.

\bibitem{zhou2020tracking}
X.~Zhou, V.~Koltun, and P.~Kr{\"a}henb{\"u}hl, ``Tracking objects as points,''
  \emph{arXiv preprint arXiv:2004.01177}, 2020.

\bibitem{chang2015shapenet}
A.~X. Chang, T.~Funkhouser, L.~Guibas, P.~Hanrahan, Q.~Huang, Z.~Li,
  S.~Savarese, M.~Savva, S.~Song, H.~Su, \emph{et~al.}, ``Shapenet: An
  information-rich 3d model repository,'' \emph{arXiv preprint
  arXiv:1512.03012}, 2015.

\bibitem{zhan2020visual}
H.~Zhan, C.~S. Weerasekera, J.-W. Bian, and I.~Reid, ``Visual odometry
  revisited: What should be learnt?'' in \emph{2020 IEEE International
  Conference on Robotics and Automation (ICRA)}.\hskip 1em plus 0.5em minus
  0.4em\relax IEEE, 2020, pp. 4203--4210.

\bibitem{Geiger2012KITTI}
A.~Geiger, P.~Lenz, and R.~Urtasun, ``Are we ready for autonomous driving? the
  kitti vision benchmark suite,'' in \emph{Conference on Computer Vision and
  Pattern Recognition (CVPR)}, 2012.

\end{thebibliography}
